\documentclass[sigconf]{acmart}

\AtBeginDocument{%
  }

\setcopyright{acmcopyright}
\copyrightyear{2018}
\acmYear{2018}
\acmDOI{XXXXXXX.XXXXXXX}

  \acmConference[DAC '24]{Make sure to enter the correct
  conference title from your rights confirmation emai}{June 23--27,
  2024}{San Francisco, CA}
\acmPrice{15.00}
\acmISBN{978-1-4503-XXXX-X/18/06}

\usepackage{enumitem}
\usepackage{xspace}
\usepackage{algorithm}
\usepackage[noend]{algpseudocode}
\usepackage{mathtools}

\newcommand{\Design}{\textit{$\mathsf{SMORE}$\xspace}}    

\DeclareMathAlphabet\mathbfcal{OMS}{cmsy}{b}{n}
\usepackage{tikz}

\let\oldbibliography\thebibliography
\renewcommand{\thebibliography}[1]{\oldbibliography{#1}
\setlength{\itemsep}{-0.12pt}} 

\usepackage{algorithm}
\usepackage{algorithmicx}
\usepackage{lipsum}
\newcommand{\RomanNumeralCaps}[1]
    {\MakeUppercase{\romannumeral #1}}
\usepackage[noend]{algpseudocode}

\algnewcommand\algorithmicforeach{\textbf{for each}}
\algdef{S}[FOR]{ForEach}[1]{\algorithmicforeach\ #1\ \algorithmicdo}

\algnewcommand{\IIf}[1]{\State\algorithmicif\ #1\  \algorithmicthen}
\algnewcommand{\EndIIf}{\unskip}
\newcommand\encircle[1]{%
  \tikz[baseline=(X.base)] 
    \node (X) [draw, shape=circle, inner sep=-1, fill=black, text=white] {{\strut #1}};%
}

\begin{document}
\title{SMORE: \underline{S}imilarity-based Hyperdi\underline mensional D\underline omain Adaptation for Multi-Senso\underline r Tim\underline{e} Series Classification}

\author{Junyao Wang, Mohammad Abdullah Al Faruque}
\email{{junyaow4, alfaruqu}@uci.edu}
\affiliation{%
  \institution{Department of Computer Science, University of California, Irvine, USA}
  \country{}
}

\begin{abstract}
Many real-world applications of the Internet of Things (IoT) employ machine learning (ML) algorithms to analyze time series information collected by interconnected sensors. 
However, \textit{distribution shift}, a fundamental challenge in data-driven ML, 
arises when a model is deployed on a data distribution different from the training data and can substantially degrade model performance. 
Additionally, increasingly sophisticated deep neural networks (DNNs) are required to capture intricate spatial and temporal dependencies in multi-sensor time series data, often exceeding the 
capabilities of today's edge devices.
In this paper, we propose $\Design$, a novel resource-efficient domain adaptation (DA) algorithm for multi-sensor time series classification, leveraging the efficient and parallel operations of hyperdimensional computing.  
$\Design$ dynamically customizes test-time models with explicit consideration of the domain context of each sample to mitigate the negative impacts of domain shifts. %
Our evaluation on a variety of multi-sensor time series classification tasks shows that $\Design$ achieves on average $1.98\%$ higher accuracy than state-of-the-art (SOTA) DNN-based 
DA algorithms with $18.81$× faster training and $4.63$× faster inference. 
\end{abstract}

\settopmatter{printacmref=false} 
\renewcommand\footnotetextcopyrightpermission[1]{} 
\maketitle

\section{Introduction}
{\begin{figure}
    \centering
    \includegraphics[width=\columnwidth]{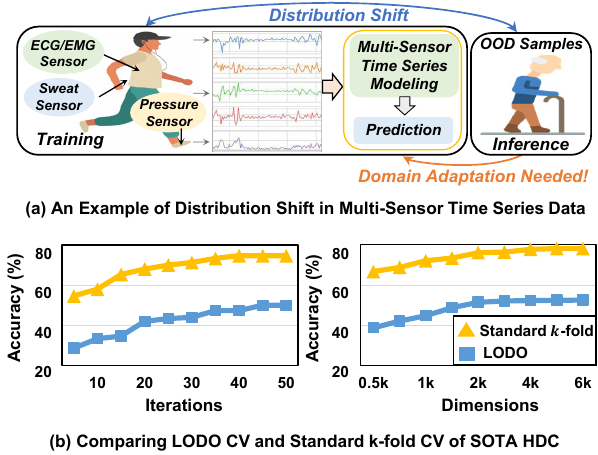}
       \vspace{-6.5mm}
    \caption{Motivation of Our Proposed $\Design$}\label{fig:motiv}
   \vspace{-3.5mm}
\end{figure}}

With the emergence of the Internet of Things (IoT), many real-world applications 
utilize heterogeneously connected sensors to collect information over the course of time, constituting multi-sensor time series data~\cite{qiao2018time}.
Machine learning (ML) algorithms, including deep neural networks (DNNs), are often employed to analyze the collected data and perform various learning tasks. 
However, \textit{distribution shift} (DS), a fundamental challenge in data-driven ML, can substantially degrade model performance. 
In particular, the excellent performance of these ML algorithms heavily relies on the critical assumption that the training and inference data come from the same distribution, while this assumption can be easily violated as out-of-distribution (OOD) scenarios are inevitable in real-world applications~\cite{wang2024rs2g, wilson2023hyperdimensional}.  
For instance, as shown in Figure \ref{fig:motiv}(a), a human activity recognition model can systematically fail when tested on individuals from different age groups or diverse demographics.

A variety of innovative \textit{domain adaptation} techniques have been proposed for deep learning (DL)~\cite{wang2020tent, zhao2018multiple, kdd, sagawadistributionally}.
However, due to their limited capacity for memorization
, DL approaches often fail to perform well on multi-sensor time series data with intricate spatial and temporal dependencies~\cite{qiao2018time, wang2023domino}. 
Although recurrent neural networks (RNNs), including long short-term memory (LSTM), have recently been  proposed to address this issue, 
these models are notably complicated and inefficient to train. 
Specifically, their complex architectures require iteratively refining millions of parameters over multiple time periods in powerful 
computing environments~\cite{qin2017dual,hochreiter1997long}. 
Considering the resource constraints of embedded devices, the massive amount of information nowadays, and the potential instabilities of IoT systems, 
a lightweight and efficient domain-adaptive learning approach 
for multi-sensor time series data is critically needed.

Brain-inspired hyperdimensional computing (HDC) has been introduced as a promising paradigm for edge ML for its 
high computational efficiency and ultra-robustness against noises
~\cite{wang2023domino,wang2023disthd, wang2023hyperdetect}. 
Additionally, 
HDC incorporates learning capability along with typical memory functions of storing/loading information, bringing unique advantages in tackling noisy time series~\cite{wang2023domino}. 
Unfortunately, existing HDCs are vulnerable to DS. 
For instance, as shown in Figure \ref{fig:motiv}(b), on the popular multi-sensor time series dataset USC-HAD~\cite{zhang2012usc}, 
SOTA HDCs converge at notably lower accuracy in leave-one-domain-out (LODO) cross-validation (CV) than in standard $k$-fold CV regardless of training iterations and model complexity. 
LODO CV develops a model with all the available data except for one domain left out for inference, while standard $k$-fold CV randomly divides data into $k$ subsets with $k-1$ subsets for training and the remaining one for evaluation. 
Such performance degradation indicates a very limited domain adaptation (DA) capability of existing HDCs. 
However, standard $k$-fold CV does not reflect real-world DS issues since the random sampling process introduces \textit{data leakage}, 
enabling the training phase to include information from all the domains 
and thus inflating model performance. 
To address this problem, 
we propose $\Design$, a novel HDC-based DA algorithm for multi-sensor time series classification. Our main contributions are listed below: 
\begin{itemize}[leftmargin = *]
    \item To the best of our knowledge, $\Design$ is the first HDC-based DA algorithm.  By explicitly considering the domain context of each sample during inference, $\Design$ provides on average $1.98\%$ higher accuracy than SOTA DL-based DA algorithms 
    on a wide range of multi-sensor time series classification tasks. 
    \item Leveraging the efficient and parallel high-dimensional operations, 
    $\Design$ provides on average $18.81\times$ faster training and $4.63\times$ faster inference than SOTA DL-based DA algorithms, providing a more efficient solution to tackle the DS challenge.
    \item We evaluate $\Design$ across multiple resource-constrained hardware devices including Raspberry Pi and NVIDIA Jetson Nano. $\Design$ demonstrates considerably lower inference latency and energy consumption than SOTA DL-based DA algorithms. 
\end{itemize}

\section{Related Works}
\subsection{Domain Adaptation}
Distribution shift (DS) arises when a model is deployed on a data distribution different from what it was trained on, posing serious robustness challenges for real-world ML applications~\cite{zhao2018multiple, wang2020tent}. 
Various innovative methodologies have been proposed to mitigate DS, and can be primarily categorized as 
\textit{domain generalizations} (DG) and \textit{domain adaptations} (DA). 
DG typically seeks to construct models by identifying domain-invariant features shared across multiple source domains~\cite{gulrajani2021in}. However, it can be challenging to extract common features when there exist multiple source domains with distinct characteristics ~\cite{dou2019domain, wang2023domino, kdd}; thus, 
existing DG approaches often fail to provide comparable high-quality results as DA~\cite{sagawadistributionally, ganin2016domain}. 
In contrast, DA generally utilizes unlabeled data in target domains to quickly adapt models trained in different source domains~\cite{wang2020tent}. 
Unfortunately, most existing DA techniques rely on multiple convolutional and fully-connected layers to train domain discriminators, requiring intensive computations and iterative refinement. 
We propose $\Design$, the first HDC-based DA algorithm, to provide a more resource-efficient DA solution for edge platforms. 

\subsection{Hyperdimensional Computing}
Several recent works have utilized HDC as a lightweight learning paradigm for time series classification~\cite{ rahimi2016hyperdimensional,moin2021wearable,wang2023robust}, 
 achieving comparable accuracy to SOTA DNNs with notably lower computational costs. 
However, existing HDCs do not consider the DS challenge. 
This can be a detrimental drawback 
as out-of-distribution (OOD) instances are inevitable during real-world deployment of ML applications. 
Hyperdimensional Feature Fusion~\cite{wilson2023hyperdimensional} identifies OOD samples by fusing outputs of several layers of a DNN model to a common
hyperdimensional space, while its backbone remains relying on resource-intensive multi-layer DNNs, and a systematic way to tackle OOD samples has yet to be proposed. 
DOMINO~\cite{wang2023domino}, a novel HDC-based DG method, constantly discards and regenerates biased dimensions representing domain-variant information; nevertheless, it requires significantly more training time to provide reasonable accuracy. 
Our proposed $\Design$ leverages efficient operations of HDC to customize test-time models for OOD samples, providing accurate predictions without causing substantial computational overhead for both training and inference. 

\section{Methodology}

\subsection{HDC Prelimnaries}\label{subsec:prelim}
Inspired by high-dimensional information representation in human brains, 
HDC maps inputs onto \textit{hyperdimensional} space as \textit{hypervectors}, each of which contains thousands of elements. 
A unique property of the hyperdimensional space is the existence of large amounts of nearly orthogonal hypervectors, enabling highly parallel and efficient operations such as similarity calculations, bundling, and binding. 
\textbf{Similarity ($\delta$):} calculation of the distance between two hypervectors. 
A common measure is cosine similarity. 
\textbf{Bundling ($+$):} element-wise addition of 
hypervectors, e.g., ${\mathcal{H}}_{bundle}= {\mathcal{H}}_1 + {\mathcal{H}}_2$, generating a hypervector with the same dimension as inputs. 
Bundling 
provides an efficient way to check the existence of a 
hypervector in a bundled set. In the previous example,  $\delta({\mathcal{H}}_{bundle},{\mathcal{H}}_1)\gg 0 $ while $\delta({\mathcal{H}}_{bundle},{\mathcal{H}}_3)\approx 0 $ (${\mathcal{H}}_3\neq {\mathcal{H}}_1, {\mathcal{H}}_2)$. Bundling models how human brains \textit{memorize} inputs. 
\textbf{Binding ($*$):} element-wise multiplication of 
two hypervectors to create another near-orthogonal hypervector, i.e. ${\mathcal{H}}_{bind} ={\mathcal{H}}_{1} * {\mathcal{H}}_{2}$ 
 where $\delta({\mathcal{H}}_{bind}, {\mathcal{H}}_{1})\approx 0$ and $\delta({\mathcal{H}}_{bind}, {\mathcal{H}}_{2})\approx 0$. 
 Due to reversibility, 
 i.e., ${\mathcal{H}}_{bind} * {\mathcal{H}}_1 = {\mathcal{H}}_2$, information from both hypervectors can be preserved. 
 Binding models how human brains \textit{connect} inputs. 
  \textbf{Permutation $(\rho)$}: a single circular shift of a hypervector by moving the value of the final dimension to the first position and shifting all other values to their next positions. 
  The permuted hypervector is nearly orthogonal to its original hypervector, i.e., $\delta(\rho \mathcal H, \mathcal H)\approx 0$. Permutation models how human brains handle \textit{sequential} inputs.


\begin{figure*}[t] 
\vspace{-0.5mm}
\centering
    \includegraphics[width=\textwidth]{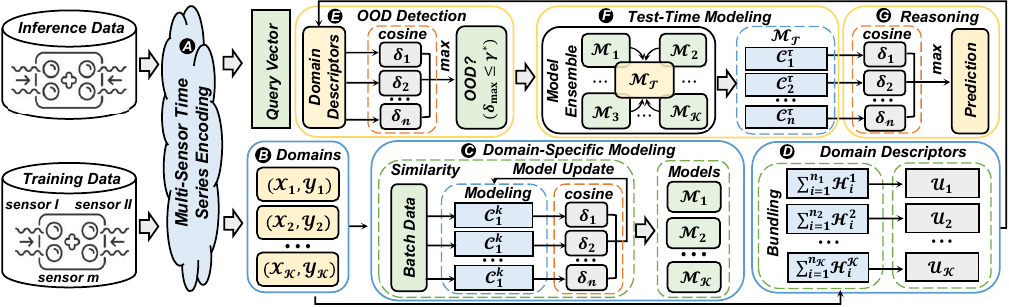}
    \vspace{-4mm}
    \caption{The Workflow of Our Proposed $\Design$}
    \label{fig:train}
\end{figure*}

\subsection{Problem Formulation}

We assume that there are $\mathcal K$ $(\mathcal K>1)$ source domains, i.e., $\mathcal D_{S}=\{\mathcal D_\mathcal S^1, \mathcal D_\mathcal S^2, \ldots, \mathcal D_\mathcal S^{\mathcal K}\}$, in the input space $\mathcal I$, and we denote the output space as $\mathcal Y$. 
$\mathcal I$ consists of time-series data from $m$ $(m\geq 1)$ interconnected sensors, i.e., $\mathcal I = \{\mathcal I_1, \mathcal I_2, \ldots, \mathcal I_m\}$. 
Our objective is to utilize training samples in $\mathcal I$ and their corresponding labels in $\mathcal Y$ to train a classification model $f:\mathcal I \rightarrow \mathcal Y$ 
to capture latent features so that we can make accurate predictions when given samples from an unseen target domain $\mathcal D_\mathcal T$.
The key challenge is that the joint distribution between source domains and target domains can be different, i.e, $\mathcal P^\mathcal S_\mathcal{(I,Y)} \neq \mathcal P^\mathcal T_\mathcal{(I,Y)}$, and thus the model $f$ can potentially fail to adapt models trained on $\mathcal D_\mathcal S$ to samples from $\mathcal D_\mathcal T$.

As shown in Figure \ref{fig:train}, our proposed $\Design$ starts with mapping training samples from the input space $\mathcal I$ to a hyperdimensional space $\mathcal X$ with an encoder $\Omega$ $(\encircle{A})$, i.e., $\mathcal X = \Omega (\mathcal I)$, that preserves the spatial and temporal dependencies in $\mathcal I$. 
We then separate the encoded data into $\mathcal K$ subsets $(\mathcal X_1, \mathcal Y_1), (\mathcal X_2, \mathcal Y_2), \ldots, (\mathcal X_\mathcal K, \mathcal Y_\mathcal K)$ $(\encircle{B})$ based on their domains. 
In the training phase, we train $\mathcal K$ \textit{domain-specific} models $\mathcal M = \{\mathcal M_1, \mathcal M_2, \ldots, \mathcal M_\mathcal K\}$ such that $\mathcal M_k: \mathcal X_k \rightarrow \mathcal Y_k$ $(1\leq k \leq \mathcal K)$ $(\encircle{C})$, 
and concurrently develop $\mathcal K$ expressive \textit{domain descriptors} $\mathcal U=\{\mathcal U_1, \mathcal U_2, \ldots, \mathcal U_\mathcal K\}$ encoding the pattern of each domain $(\encircle{D})$. 
When given a inference sample $\mathcal I_{\mathcal T}$ from the target domain $\mathcal D_\mathcal T$, we map $\mathcal I_{\mathcal T}$ to hyperdimensional space with the same encoder $\Omega$ used for training, 
and identify OOD samples ($\encircle E$) with a binary classifier $\Phi$ utilizing the domain descriptors $\mathcal U$, i.e., $\Phi (\Omega(\mathcal I_{\mathcal T}), \mathcal U)$. 
Finally, we construct a \textit{test-time model} $\mathcal M_\mathcal T$ based on domain-specific models $\mathcal M$ ($\encircle F$) and whether the sample is OOD to make inferences ($\encircle G$) with explicit consideration of the domain context of $\mathcal I_\mathcal T$, i.e., $\hat{\mathcal Y_\mathcal I}=\mathcal M_\mathcal T(\Phi(\Omega(\mathcal I_{\mathcal T}), \mathcal U), \mathcal M)$.



\subsection{Multi-Sensor Time Series Data Encoding}\label{subsec:tsencode}
{\begin{figure}
\vspace{-2mm}
    \centering
    \includegraphics[width=\columnwidth]{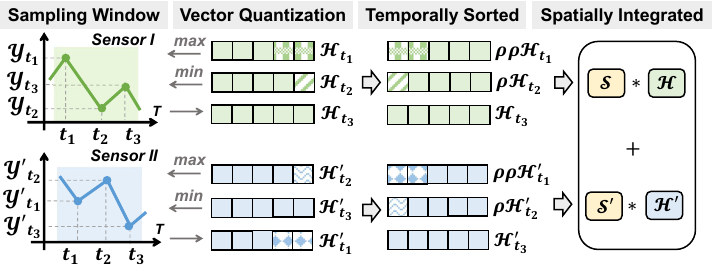}
    \caption{HDC Encoding for Multi-Sensor Time Series Data}\label{fig:tsencode}
   \vspace{-4mm}
\end{figure}}
We employ the encoding techniques in Figure \ref{fig:tsencode} to capture and preserve spatial and temporal dependencies in multi-sensor time series data when mapping low-dimension inputs to hyperdimensional space. 
We sample time series data in $n$-gram windows; in each sample window, the signal values ($y$-axis) store the information and the time ($x$-axis) represents the temporal sequence. 
We first assign random hypervectors $\mathcal H_{max}$ and $\mathcal H_{min}$ to represent the maximum and minimum signal values. 
We then perform vector quantization to values between the maximum and minimum values to generate vectors with a spectrum of similarity to $\mathcal H_{max}$ and $\mathcal H_{min}.$ 
For instance, in Figure \ref{fig:tsencode}, Sensor \RomanNumeralCaps{1} and Sensor \RomanNumeralCaps{2} follow a time series in trigram. 
Sensor \RomanNumeralCaps{1} has the maximum value at $t_1$ and the minimum value at $t_2$, and thus we assign randomly generated hypervectors $\mathcal H_{t_1}$ and $\mathcal H_{t_2}$ to $y_{t_1}$ and $y_{t_2}$, respectively. 
Similarly, we assign randomly generated hypervectors $\mathcal H'_{t_2}$ and $\mathcal H'_{t_3}$ to $y'_{t_2}$ and $y'_{t_3}$ in Sensor \RomanNumeralCaps{2}. We then assign hypervectors to $y_{t_3}$ in Sensor \RomanNumeralCaps{1} and value at $y'_{t_1}$ in Sensor \RomanNumeralCaps{2} with vector quantization; mathematically, 
\begin{align*}
    &\mathcal H_{t_3} = \mathcal H_{t_2}+\frac{y_{t_3}-y_{t_2}}{y_{t_1}-y_{t_2}}\cdot (\mathcal H_{t_1} - \mathcal H_{t_2})\\
    & \mathcal H'_{t_1} = \mathcal H'_{t_3}+\frac{y'_{t_1}-y'_{t_3}}{y'_{t_2}-y'_{t_3}}\cdot (\mathcal H'_{t_2} - \mathcal H'_{t_3}).
\end{align*}
We represent the temporal sequence of data with the permutation operation in section \ref{subsec:prelim}. 
For Sensor \RomanNumeralCaps{1} and Sensor \RomanNumeralCaps{2} in Figure \ref{fig:tsencode}, we perform rotation shift ($\rho$) twice to $\mathcal H_{t_1}$ and $\mathcal H'_{t_1}$, once to $\mathcal H_{t_2}$ and $\mathcal H'_{t_2}$, and keep $\mathcal H_{t_3}$ and $\mathcal H'_{t_3}$  the same. 
We bind data samples in one sampling widow by calculating $\mathcal H = \rho\rho\mathcal H_{t_1} * \rho\mathcal H_{t_2} * \mathcal H_{t_3}$ and $\mathcal H' = \rho\rho\mathcal H'_{t_1} * \rho\mathcal H'_{t_2} * \mathcal H'_{t_3}$. Finally, to spatially integrate data from multiple sensors, we generate a random signature hypervector for each sensor and bind information as $\sum_{i=1}^m [\mathcal G_i * \mathcal H_i]$, 
where $\mathcal G_i$ denotes the signature hypervector for sensor $i$, $\mathcal H_i$ is the hypervector containing overall information from sensor $i$, and $m$ denotes the total number of sensors. 
For our example in Figure \ref{fig:tsencode}, we combine information from Sensor \RomanNumeralCaps{1} and Sensor \RomanNumeralCaps{2} by randomly generating signature hypervectors $\mathcal G$ and $\mathcal G'$ and calculating $(\mathcal G * \mathcal H) + (\mathcal G' * \mathcal H')$.

\subsection{Domain-Specific Modeling}\label{subsec:dsmodel}
As shown in Figure \ref{fig:train}, after mapping data to hyperdimensional space (\encircle{A}), 
$\Design$ separates training samples into $\mathcal K$ subsets 
based on their domains (\encircle{B}), where $\mathcal K$ represents the total number of domains.. 
We then employ a highly efficient HDC algorithm to calculate a \textit{domain-specific} model for every domain (\encircle{C}), i.e., $\mathcal M = \{\mathcal M_1, \mathcal M_2, \ldots, \mathcal M_\mathcal K\}$.
Our approach aims to provide high-quality results with fast convergence by identifying common patterns during training and eliminating model saturations. 
We bundle data points by scaling a proper weight to each of them depending on how much new information is added to class hypervectors. 
In particular, each domain-specific model $\mathcal M_k (1\leq k \leq \mathcal K)$ consists of $n$ class hypervectors $\mathcal C_1^k, \mathcal C_2^k, \ldots, \mathcal C_n^k$ ($n =$ the number of classes), each of which encodes the pattern of a class. 
A new sample $\mathcal H$ in domain $\mathcal D_\mathcal S^k$ updates model $\mathcal M_k$ based on its cosine similarities, denoted as $\delta(\mathcal H, \cdot)$, with all the class hypervectors in $\mathcal M_k$, 
i.e., 
\begin{equation}\label{eqn:cos}
    \delta( {\mathcal H}, {\mathcal C_{t}^k})=\frac{ {\mathcal H} \cdot {\mathcal C_{t}^k}}{\| {\mathcal H}\|\cdot \| {\mathcal C_{t}^k}\|}=\frac{ {\mathcal H}}{\| {\mathcal H}\|}\cdot \frac{{\mathcal C_{t}^k}}{\| {\mathcal C_{t}^k}\|}\propto {\mathcal H}\cdot {\textit{Norm}({\mathcal C_{t}^k})}
\end{equation}
where 
$1\leq t\leq n$. 
If ${\mathcal H}$ has the highest cosine similarity with the class hypervector $\mathcal C_i^k$ while its true label $\mathcal C_j^k$ $(1\leq i, j\leq n)$, 
the domain-specific model $\mathcal M_k$ updates as 
\begin{equation} \label{eqn:update}
    \begin{split}
        &\mathcal C_j^k \leftarrow \mathcal C_j^k + \eta\cdot\Big(1-\delta{\big(\mathcal H, \mathcal C_j^k\big)}\Big)\times{\mathcal H }\\
        & \mathcal C_i^k \leftarrow \mathcal C_i^k - \eta\cdot\Big(1-\delta{\big(\mathcal H, \mathcal C_i^k\big)}\Big)\times{\mathcal H},
    \end{split} 
\end{equation}
where $\eta$ denotes a learning rate. 
A large $\delta(\mathcal H, \cdot)$ indicates the input data point is marginally mismatched or already exists in the model, and the model is updated by adding a very small portion of the encoded vector ($1-\delta(\mathcal H, \cdot) \approx 0$). 
In contrast, a small $\delta(\mathcal H, \cdot)$ indicates a noticeably new pattern and 
updates the model with a large factor ($1-\delta(\mathcal H, \cdot) \approx 1$). 
Our learning algorithm provides a higher chance for non-common patterns to be properly included in the model. 

\vspace{-0.2mm}
\subsection{Out-of-Distribution Detection} \label{subsec:ood}
\subsubsection{Domain Descriptors}
In parallel to domain-specific modeling, as shown in Figure \ref{fig:train}, we concurrently construct \textit{domain descriptors} $(\encircle D)$ $\mathcal U = \{\mathcal U_1, \mathcal U_2, \ldots, \mathcal U_\mathcal K\}$ to encode the distinct pattern of each domain.  
Specifically,  for each domain $\mathcal D_\mathcal S ^k$ $(1\leq k \leq \mathcal K)$, we utilize the bundling operation to combine the hypervector within the domain, i.e., $\{\mathcal H_1^k, \mathcal H_2^k,  \ldots, \mathcal H_{n_k}^k\}$, and construct an expressive descriptor $\mathcal U_k$. Mathematically, $\mathcal U_k = \sum_i^{n_k} \mathcal H_i^k$. 
Given the property of the bundling operation (explained in section \ref{subsec:prelim}), $\mathcal U_k$ is cosine-similar to all the samples $\mathcal H_i^k$ ($1\leq i \leq n_k$) within domain $\mathcal D_\mathcal S ^k$ as they contribute to the bundling process, and dissimilar to all the samples that are not part of the bundle, i.e., not 
in domain $\mathcal D_\mathcal S ^k$.

\subsubsection{Out-of-Distribution Detection} 
As shown in Figure \ref{fig:train}, a key component of our inference phase is the detection of OOD samples (\encircle E). We start with mapping the testing sample to hyperdimensional space with the same encoding technique as the training phase to obtain a query vector $\mathcal Q$ (\encircle A). Then, as detailed in Algorithm \ref{alg:infer}, we calculate the cosine similarity score of the query vector $\mathcal Q$ to each domain descriptor $\mathcal U_1, \mathcal U_2, \ldots, \mathcal U_\mathcal K$ (line \ref{alg:line:maxdelta}). 
A testing sample is identified as OOD if its pattern is substantially different from all the source domains. 
Therefore, 
when the cosine similarity between $\mathcal Q$ and its most similar domain, i.e., $\max\big\{\delta(\mathcal Q, \mathcal U_1), \delta(\mathcal Q, \mathcal U_2), \ldots, \delta(\mathcal Q, \mathcal U_\mathcal K)\big\}$, is smaller than a threshold $\mathcal \delta^*$, 
we consider $\mathcal Q$ as an OOD sample (line \ref{alg:line:ood_criteria}).
Here $\delta^*$ is a tunable parameter and we analyze the impact of $\delta^*$ in section \ref{subsec:param}. 
We then dynamically construct test-time models (\encircle F) based on whether the sample is OOD (section \ref{subsec:modeling}). 
\subsection{Adaptive Test-Time Modeling}\label{subsec:modeling}
\subsubsection{Inference for OOD Samples}\label{subsubsec:ood_model}
For each testing sample identified as OOD, we dynamically adapt domain-specific models $\mathcal M = \{\mathcal M_1, \mathcal M_2, \ldots, \mathcal M_k\}$ to customize a test-time model $\mathcal M_\mathcal T$ that best fits its domain context and thereby provides an accurate prediction. 
As detailed in Algorithm \ref{alg:infer}, for an OOD sample $\mathcal Q$, we ensemble each domain-specific model based on how similar $\mathcal Q$ is to the domain (line \ref{alg:line:ensemble}). 
Specifically, let $\delta(\mathcal Q, \mathcal U_1)$, $\delta(\mathcal Q, \mathcal U_2)$, $\dots$, $\delta(\mathcal Q, \mathcal U_\mathcal K)$ denote the cosine similarity between $\mathcal Q$ and each domain descriptor $\mathcal U_1, \mathcal U_2, \ldots, \mathcal U_\mathcal K$, 
we construct the test-time model $\mathcal M_\mathcal T$ for $\mathcal Q$ as
\vspace{-0.5mm}
\begin{equation}
    \mathcal M_{\mathcal T} = \delta(\mathcal Q, \mathcal U_1) \cdot \mathcal M_1 + \delta(\mathcal Q, \mathcal U_2) \cdot \mathcal M_2 + \ldots + \delta(\mathcal Q, \mathcal U_\mathcal K) \cdot \mathcal M_\mathcal K.
\end{equation}
Specifically, $\mathcal M_\mathcal T$ is of the same shape as $\mathcal M_1, \mathcal M_2, \ldots, \mathcal M_k$; it consists of $n$ class hypervectors ($n=$ number of classes), denoted as $\mathcal C_1^{{\mathcal T}}, \mathcal C_2^{{\mathcal T}}, \ldots, \mathcal C_n^{{\mathcal T}}$, formulated with explicit consideration of the domain context of $\mathcal Q$.
We then compute the cosine similarity between $\mathcal Q$ and each of these class hypervectors, and assign $\mathcal Q$ to the class to which it achieves the highest similarity score (line \ref{alg:line:pred}). 
\vspace{-2mm}
\subsubsection{Inference for In-Distribution Samples}
{We predict the label of an in-distribution testing sample, i.e., a non-OOD sample, leveraging domain-specific models of the domains to which the sample is highly similar.  
As demonstrated in Algorithm \ref{alg:infer}, for all the domains $\mathcal D_\mathcal S^i$ ($1\leq i\leq \mathcal K$) where the encoded query vector $\mathcal Q$ achieves a similarity score higher than $\delta^*$, we ensemble their corresponding domain-specific models incorporating a weight of their cosine similarity score with $\mathcal Q$ to formulate the test-time model $\mathcal M_\mathcal T$ (line \ref{alg:line:in_criteria} - \ref{alg:line:in_ensemble}). Similar to section \ref{subsubsec:ood_model}, $\mathcal M_\mathcal T$ consists of $n$ class hypervectors, and $\mathcal Q$ is assigned to the class where it obtained the highest similarity score (line \ref{alg:line:pred}).  
Note that, unlike the inference for OOD samples where we ensemble all the domain-specific models to enhance performance, the test-time model $\mathcal M_\mathcal T$ for an in-distribution sample does not consider domain-specific models of the domains where $\mathcal Q$ show a minor similarity score lower than $\delta^*$. 
In particular, if a testing sample $\mathcal Q$ does not show high similarity to any of the source domains, we include information from all the domains to construct a sufficiently comprehensive test-time model to mitigate the negative impacts of distribution shift. 
In contrast, when $\mathcal Q$ exhibits considerable similarity to certain domains, adding information from other domains is comparable to introducing noises and can potentially mislead the classification and degrade model performance. }


\begin{algorithm}[t]
\small
  \caption{Domain Adaptive HDC Inference}\label{alg:infer}
  \begin{algorithmic}[1]{}
\Require{An encoded testing samples $\mathcal Q$, a domain classifier with $\mathcal K$ class hypervectors $\mathcal U_1,\mathcal U_2, \ldots,\mathcal U_\mathcal K$, a threshold for OOD detection $\delta^*$, $\mathcal K$ domain-specific models $\mathcal M_1, \mathcal M_2, \ldots ,\mathcal M_{\mathcal K}$}
    
    \Ensure{A predicted label $\mathcal P$ for $\mathcal Q$.} 
    \State $\delta_{max} = \max\{\delta(\mathcal Q, \mathcal U_1), \delta(\mathcal Q, \mathcal U_2)\ldots, \delta(\mathcal Q, \mathcal U_\mathcal K)\}$ \label{alg:line:maxdelta} \Comment{OOD detection}
    \If{$\delta_{max}<\delta^*$} \Comment{$\mathcal Q$ is considered OOD} \label{alg:line:ood_criteria}
        \State $\mathcal M_\mathcal T \leftarrow \sum_{i=1}^{\mathcal K} \delta(\mathcal Q, \mathcal U_i)\cdot \mathcal M_i$ \Comment{model ensembling} \label{alg:line:ensemble}
    \Else \Comment{$\mathcal Q$ is not OOD}
       \ForAll{$\delta(\mathcal Q, \mathcal U_i) \geq \delta^*$} \Comment{$1\leq i \leq k$} \label{alg:line:in_criteria}
            \State{$\mathcal M_\mathcal T \leftarrow \sum_{i=1}^\mathcal K \delta(\mathcal Q, \mathcal U_i)\cdot \mathcal M_i$} \Comment{partial model ensembling} \label{alg:line:in_ensemble}
        \EndFor
    \EndIf
    \State{$\mathcal P \leftarrow \arg\max_{\mathcal C_i^\mathcal T}\{\delta(\mathcal Q, \mathcal C_1^{\mathcal T}), \delta(\mathcal Q, \mathcal C_2^{\mathcal T})\ldots, \delta(Q,\mathcal C_n^{\mathcal T})$\}} 
 \label{alg:line:pred}
    \State \Return $\mathcal P$
  \end{algorithmic}
\end{algorithm}
\setlength{\textfloatsep}{8pt}



\section{Experimental Evaluations}

\begin{table}[!t]
\vspace{-2.5mm}
\footnotesize
\centering\caption{Detailed Breakdowns of Datasets}\vspace{-4mm}
\centering{($\mathcal N$: number of data samples)} 
\vspace{-0.5mm}
\begin{center}
\begin{tabular}{cc|cc|cc} \toprule
\multicolumn{2}{c}{DSADS~\cite{barshan2014recognizing}}& \multicolumn{2}{c}{USC-HAD~\cite{zhang2012usc}} &\multicolumn{2}{c}{PAMAP2~\cite{reiss2012introducing}} \\
\midrule
 Domains & $\mathcal N$  & Domains  & $\mathcal N$ &  Domains  &  $\mathcal N$\\
\midrule
Domain 1 & 2,280 & Domain 1 &  8,945	 &	 Domain 1 & 5,636\\
Domain 2 & 2,280 & Domain 2  &  8,754 &  Domain 2  & 5,591\\
Domain 3 & 2,280  & Domain 3 & 8,534 &  Domain 3 & 5,806\\
Domain 4 & 2,280  & Domain 4 & 8,867  &  Domain 4 & 5,660\\
 & & Domain 5& 8,274& &\\ \midrule
Total & 9,120 & Total & 43,374 & Total &22,693\\
\bottomrule
\end{tabular}
\label{tb:detail_data}
\end{center}
\end{table}

\subsection{Experimental Setup}
We evaluate $\Design$ on widely-used multi-sensor time series datasets DSADS~\cite{barshan2014recognizing}, USC-HAD~\cite{zhang2012usc}, PAMAP2~\cite{reiss2012introducing}.
Domains are defined by subject grouping chosen based on subject ID from low to high.
The data size of each domain in each dataset is demonstrated in TABLE \ref{tb:detail_data}.
We compare $\Design$ with (\romannumeral 1) two SOTA CNN-based DA algorithms: {TENT~\cite{wang2020tent} and multisource domain adversarial networks (MDANs)~\cite{zhao2018multiple}}, and (\romannumeral 2) two HDC algorithms: the SOTA HDC not considering distribution shifts~\cite{hernandez2021onlinehd} (BaselineHD) and DOMINO~\cite{wang2023domino}, a recently proposed HDC-based domain generalization framework.
 The CNN-based DA algorithms are trained with TensorFlow, and we apply the common practice of grid search to identify the best hyper-parameters for each model.
Since DOMINO involves dimension regeneration in every training iteration, for fairness, we initiate it with dimension $d^* = 1k$ and make its total dimensionalities, i.e., the sum of its initial dimension and all the regenerated dimensions throughout retraining, the same as $\Design$ and BaselineHD ($d = 8k$). 
Our evaluations include leave-one-domain-out (LODO) performance, learning efficiency on both server CPU and resource-constrained devices, and scalability using different sizes of data.

{\begin{figure}
    \centering
    \includegraphics[width=\columnwidth]{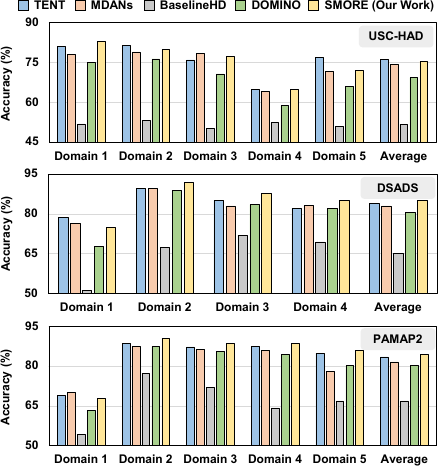}
       \vspace{-6mm}
    \caption{Comparing LODO Accuracy of $\Design$ and CNN-based Domain Adaptation Algorithms}\label{fig:acc}
  \vspace{-1mm}
\end{figure}}

\subsubsection{Platforms}
To evaluate the performance of $\Design$ on both high-performance computing environments and resource-limited edge devices, we include results from the following platforms:
\begin{itemize}[leftmargin=*]
  \item \textbf{Server CPU}: Intel Xeon Silver 4310 CPU (12-core, 24-thread, 2.10 GHz), 96 GB DDR4 memory, Ubuntu 20.04, Python 3.8.10, PyTorch 1.12.1, TDP 120 W.
  \item \textbf{Embedded CPU}: Raspberry Pi 3 Model 3+ (quad-core ARM A53 @1.4GHz), 1 GB LPDDR2 memory, Debian 11, Python 3.9.2, PyTorch 1.13.1, TDP 5 W.
  \item \textbf{Embedded GPU}: Jetson Nano (quad-core ARM A57 @1.43 GHz, 128-core Maxwell GPU), 4 GB LPDDR4 memory, Python 3.8.10, PyTorch 1.13.0, CUDA 10.2,  TDP 10 W. 
  \end{itemize}

\subsubsection{Data Preprocessing}\label{sec:eval}
We describe the data processing 
for each dataset primarily 
on data segmentation and domain labeling.
\begin{itemize}[leftmargin=*]
\item\textbf{DSADS}\cite{barshan2014recognizing}: The Daily and Sports Activities Dataset (DSADS) includes 19 activities performed by 8 subjects. 
Each data segment is a non-overlapping five-second window sampled at 25Hz.
Four domains are formed with two subjects each.

\item\textbf{USC-HAD}~\cite{zhang2012usc}: The USC human activity dataset (USC-HAD) includes 12 activities performed by 14 subjects.
Each data segment is a 1.26-second window sampled at 100Hz with 50\% overlap.
Five domains are formed with three subjects each.

\item\textbf{PAMAP2}~\cite{reiss2012introducing}: The Physical Activity Monitoring (PAMAP2) dataset includes 18 activities performed by 9 subjects.
Each data segment is a 1.27-second window sampled at 100Hz with 50\% overlap.
Four domains, excluding subject nine, are formed with two subjects each.
 \end{itemize}

\subsection{Accuracy}
The accuracy of the LODO classification is demonstrated in Figure \ref{fig:acc}. The accuracy of Domain $k$ indicates that the model is trained with data from all the other domains and evaluated with data from Domain $k$. 
This accuracy score serves as an indicator of the model's domain adaptation capability when confronted with unseen data from unseen distributions. 
$\Design$ achieves comparable performance to TENT and on average $1.98\%$ higher accuracy than MDANs. 
Additionally, $\Design$ provides $20.25\%$ higher accuracy than BaselineHD, demonstrating that $\Design$ successfully adapts trained models to fit samples from diverse domains. Additionally, $\Design$ provides on average $4.56\%$ higher accuracy than DOMINO, indicating DA can more effectively address the domain distribution (DS) challenge in noisy multi-sensor time series data compared to DG.

 \subsubsection{Impact of Hyperparameter $\delta^*$} \label{subsec:param}
The impact of $\delta^*$ on classification results is demonstrated in Figure \ref{fig:dse}, where we evaluate $\Design$ on the dataset USC-HAD as an example. 
$\Design$ achieved its best performance when $\delta^*$ is around 0.65. 
Small values of $\delta^*$ are more likely to detect in-distribution samples as OOD so that the test-time models would include noises from the domains where the sample only has a minimal similarity score, thereby causing notable performance degradation.
On the other hand, large values of $\delta^*$  are more likely to consider OOD samples in-distribution. Consequently, the test-time model only includes domain-specific models of the domains that exhibit limited similarity to the given sample; therefore, the ensembled test-time model is unlikely to be sufficiently comprehensive to provide accurate predictions.

{\begin{figure}
    \centering
    \includegraphics[width=\columnwidth]{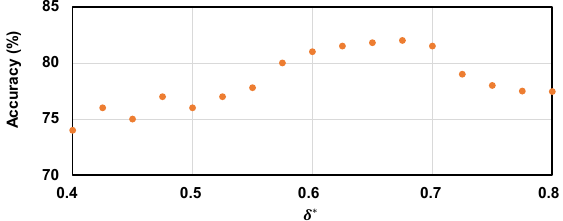}
       \vspace{-6mm}
    \caption{Imapct of $\delta^*$ on Model Performance}\label{fig:dse}
  \vspace{-1mm}
\end{figure}}


\subsection{Efficiency}
{\begin{figure}
    \centering
    \includegraphics[width=\columnwidth]{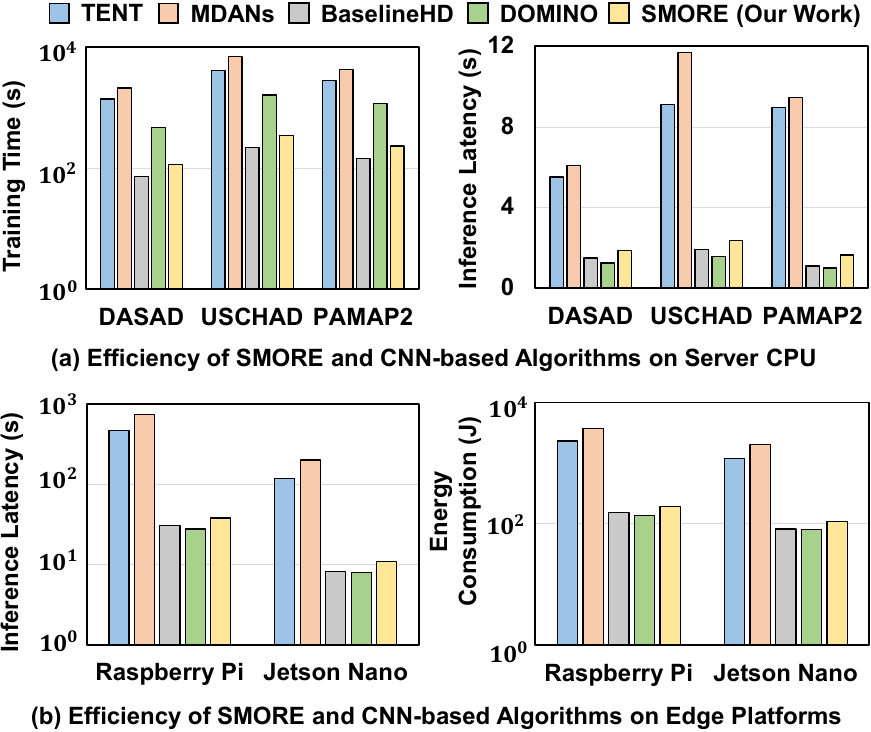}
       \vspace{-6mm}
    \caption{Efficiency of $\Design$ and CNN-based DA Algorithms}\label{fig:eff}
 \vspace{-1.5mm}
\end{figure}}

{\begin{figure}
    \centering
    \includegraphics[width=\columnwidth]{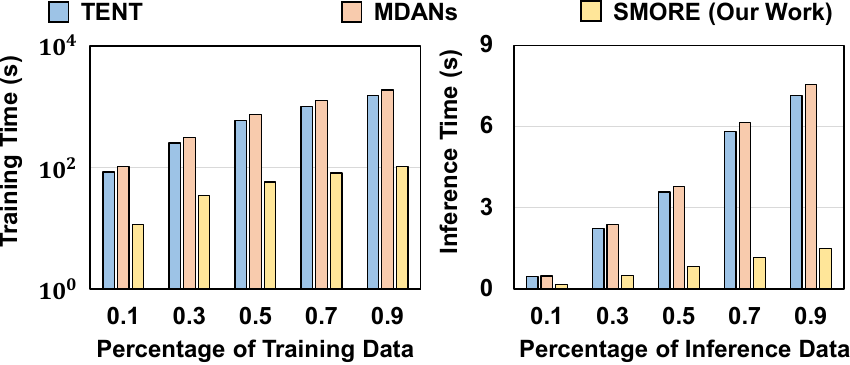}
        \vspace{-6mm}
    \caption{Comparing Scalability Using Different Size of Data}\label{fig:scale}
   \vspace{-1mm}
\end{figure}}
\subsubsection{Efficiency on Server CPU}
For each dataset, each domain consists of roughly similar amounts of data as detailed in TABLE \ref{tb:detail_data}; thus, we show the average runtime of training and inference for all the domains. 
As demonstrated in Figure \ref{fig:eff}(a), $\Design$ exhibits on average $11.64\times$ faster training than TENT and $18.81\times$ faster training than MDANs. 
Additionally, $\Design$ delivers $4.07\times$ faster inference than TENT and $4.63\times$ faster inference than MDANs. 
Such notably higher learning efficiency is thanks to the highly parallel matrix operations on hyperdimensional space. 
Additionally, $\Design$ provides on average $5.84\times$ faster training compared to DOMINO. In particular, DOMINO achieves domain generalization by iteratively identifying and regenerating domain-variant dimensions and thus requires notably more retraining iterations to provide reasonable performance. 
On the other hand, during each retraining iteration, DOMINO only keeps dimensions playing the most positive role in the classification task, and thus it eventually arrives at a model with compressed dimensionality and provides a slightly higher inference efficiency. 
Furthermore, compared to BaselineHD, $\Design$ provides significantly higher accuracy (demonstrated in Fig. \ref{fig:acc}) without substantially increasing both the training and inference runtimes.

\subsubsection{Efficiency on Embedded Platforms}
To further understand the performance of $\Design$ on resource-constrained computing platforms, we evaluate the efficiency of $\Design$, TENT, MDANs, and BaselineHD using a Raspberry Pi 3 Model B+ board and an NVIDIA Jetson Nano board. 
Both platforms have very limited memory and CPU cores (and GPU cores for Jetson Nano). 
Figure \ref{fig:eff}(b) shows the average inference latency for each algorithm processing each domain in the PAMAP2 dataset. 
{
On Raspberry Pi, $\Design$ provides on average$14.82\times$ speedups compared to TENT and $19.29\times$ speedups compared to MDANs in inference. 
On Jetson Nano, $\Design$ delivers $13.22\times$ faster inference than TENT and $17.59\times$ faster inference than MDANs. }
Additionally, $\Design$ exhibits significantly less energy consumption, providing a more resource-efficient domain adaptation solution for the distribution shift challenge on edge devices.  
\subsection{Scalability}
We evaluate the scalability of $\Design$ and SOTA CNN-based DA algorithms using various training data sizes (percentages of the full dataset) of the PAMAP2 dataset. As demonstrated in Figure \ref{fig:scale}, with increasing the training data size, $\Design$ maintains high efficiency in both training and inference with a sub-linear growth in execution time. In contrast, the training and inference time of CNN-based DA algorithms increases considerably faster than $\Design$.  This positions $\Design$ as an efficient and scalable DA solution for both high-performance and resource-constrained computing platforms.

\section{Conclusions}
We propose $\Design$, a novel HDC-based domain adaptive algorithm that dynamically customizes test-time models with explicit consideration of the domain context of each sample and thereby provides accurate predictions. $\Design$ outperforms SOTA CNN-based DA algorithms in both accuracy and training and inference efficiency, 
making it an ideal solution to address the DS challenge. 


\section{Acknowledgement}
This work was partially supported by the National Science Foundation (NSF) under award CCF-2140154.

\bibliographystyle{unsrt}
\bibliography{ref}

\begin{thebibliography}{10}

\bibitem{qiao2018time}
Huihui Qiao et~al.
\newblock A time-distributed spatiotemporal feature learning method for machine health monitoring with multi-sensor time series.
\newblock {\em Sensors}, 2018.

\bibitem{wang2024rs2g}
Junyao Wang et~al.
\newblock Rs2g: Data-driven scene-graph extraction and embedding for robust autonomous perception and scenario understanding.
\newblock In {\em Proceedings of the IEEE/CVF Winter Conference on Applications of Computer Vision}, 2024.

\bibitem{wilson2023hyperdimensional}
Samuel Wilson et~al.
\newblock Hyperdimensional feature fusion for out-of-distribution detection.
\newblock In {\em Proceedings of the IEEE/CVF Winter Conference on Applications of Computer Vision}, 2023.

\bibitem{wang2020tent}
Dequan Wang et~al.
\newblock Tent: Fully test-time adaptation by entropy minimization.
\newblock In {\em International Conference on Learning Representations}, 2020.

\bibitem{zhao2018multiple}
Han Zhao et~al.
\newblock Multiple source domain adaptation with adversarial learning.
\newblock In {\em 6th International Conference on Learning Representations, ICLR 2018}, 2018.

\bibitem{kdd}
Xin Qin et~al.
\newblock Generalizable low-resource activity recognition with diverse and discriminative representation learning.
\newblock In {\em Proceedings of the 29th ACM SIGKDD Conference on Knowledge Discovery and Data Mining}, page 1943–1953, 2023.

\bibitem{sagawadistributionally}
Shiori Sagawa et~al.
\newblock Distributionally robust neural networks.
\newblock In {\em International Conference on Learning Representations}, 2019.

\bibitem{wang2023domino}
Junyao Wang, Luke Chen, and Mohammad~Abdullah Al~Faruque.
\newblock Domino: Domain-invariant hyperdimensional classification for multi-sensor time series data.
\newblock In {\em 2023 IEEE/ACM International Conference on Computer Aided Design (ICCAD)}, pages 1--9. IEEE, 2023.

\bibitem{qin2017dual}
Yao Qin et~al.
\newblock A dual-stage attention-based recurrent neural network for time series prediction.
\newblock In {\em Proceedings of the 26th International Joint Conference on Artificial Intelligence}, 2017.

\bibitem{hochreiter1997long}
Sepp Hochreiter et~al.
\newblock Long short-term memory.
\newblock {\em Neural computation}, 1997.

\bibitem{wang2023disthd}
Junyao Wang et~al.
\newblock Disthd: A learner-aware dynamic encoding method for hyperdimensional classification.
\newblock {\em arXiv preprint arXiv:2304.05503}, 2023.

\bibitem{wang2023hyperdetect}
Junyao Wang et~al.
\newblock Hyperdetect: A real-time hyperdimensional solution for intrusion detection in iot networks.
\newblock {\em IEEE Internet of Things Journal}, 2023.

\bibitem{zhang2012usc}
Mi~Zhang et~al.
\newblock Usc-had: A daily activity dataset for ubiquitous activity recognition using wearable sensors.
\newblock In {\em Proceedings of the 2012 ACM conference on ubiquitous computing}, 2012.

\bibitem{gulrajani2021in}
Ishaan Gulrajani et~al.
\newblock In search of lost domain generalization.
\newblock In {\em International Conference on Learning Representations}, 2021.

\bibitem{dou2019domain}
Qi~Dou et~al.
\newblock Domain generalization via model-agnostic learning of semantic features.
\newblock {\em Advances in Neural Information Processing Systems}, 2019.

\bibitem{ganin2016domain}
Yaroslav Ganin et~al.
\newblock Domain-adversarial training of neural networks.
\newblock {\em The journal of machine learning research}, 2016.

\bibitem{rahimi2016hyperdimensional}
Abbas Rahimi et~al.
\newblock Hyperdimensional biosignal processing: A case study for emg-based hand gesture recognition.
\newblock In {\em ICRC}. IEEE, 2016.

\bibitem{moin2021wearable}
Ali Moin et~al.
\newblock A wearable biosensing system with in-sensor adaptive machine learning for hand gesture recognition.
\newblock {\em Nature Electronics}, 2021.

\bibitem{wang2023robust}
Junyao Wang et~al.
\newblock Robust and scalable hyperdimensional computing with brain-like neural adaptations.
\newblock {\em arXiv preprint arXiv:2311.07705}, 2023.

\bibitem{barshan2014recognizing}
Billur Barshan et~al.
\newblock Recognizing daily and sports activities in two open source machine learning environments using body-worn sensor units.
\newblock {\em The Computer Journal}, 2014.

\bibitem{reiss2012introducing}
Attila Reiss et~al.
\newblock Introducing a new benchmarked dataset for activity monitoring.
\newblock In {\em 16th international symposium on wearable computers}. IEEE, 2012.

\bibitem{hernandez2021onlinehd}
Alejandro Hern{\'a}ndez-Cano et~al.
\newblock Onlinehd: Robust, efficient, and single-pass online learning using hyperdimensional system.
\newblock In {\em DATE}. IEEE, 2021.

\end{thebibliography}

\end{document}